\documentclass{article}

\usepackage[nonatbib,preprint]{neurips_2025}

\usepackage{amsmath}

\usepackage[utf8]{inputenc} %
\usepackage[T1]{fontenc}    %
\usepackage{hyperref}       %
\usepackage{url}            %
\usepackage{booktabs}       %
\usepackage{amsfonts}       %
\usepackage{nicefrac}       %
\usepackage{microtype}      %
\usepackage{xcolor}         %
\usepackage{graphicx}
\usepackage[numbers]{natbib}
\usepackage{listings}
\usepackage{wrapfig}

\usepackage{hyperref}
\usepackage{cleveref}
\usepackage{multirow}
\usepackage{xcolor}

\definecolor{codegreen}{rgb}{0,0.6,0}
\definecolor{codegray}{rgb}{0.5,0.5,0.5}
\definecolor{codepurple}{rgb}{0.58,0,0.82}
\definecolor{backcolour}{rgb}{0.95,0.95,0.95}

\lstdefinestyle{mystyle}{
    backgroundcolor=\color{backcolour},   
    commentstyle=\color{codegreen},
    keywordstyle=\color{magenta},
    numberstyle=\tiny\color{codegray},
    stringstyle=\color{codepurple},
    basicstyle=\ttfamily\footnotesize,
    breakatwhitespace=false,         
    breaklines=true,                 
    captionpos=b,                    
    keepspaces=true,                 
    numbers=left,                    
    numbersep=5pt,                  
    showspaces=false,                
    showstringspaces=false,
    showtabs=false,                  
    tabsize=2
}

\usepackage{xcolor}

\definecolor{TeseoColor}{rgb}{0.15, 0.68, 0.38}
\definecolor{DanieleColor}{rgb}{1, 0, 1}
\definecolor{NafisehColor}{rgb}{0.65, 0.5, 0.65}
\definecolor{SaiColor}{rgb}{0.65, 0.9, 0.65}

\newcommand{\RR}{\mathbb{R}}

\definecolor{Violet}{rgb}{.6, .35, .7}
\definecolor{Pink}{rgb}{1, 0, 1}
\definecolor{Yellow}{rgb}{1, 0.65, 0}

\title{Better STEP, a format and dataset for boundary representation}

\author{%
  Nafiseh Izadyar \\
  Department of Computer Science\\
  University of Victoria\\
  \And
  Sai Chandra Madduri\\
  Department of Computer Science\\
  University of Victoria\\
  \And
  Teseo Schneider\\
  Department of Computer Science\\
  University of Victoria\\
}

\begin{document}

\maketitle

\begin{figure}[h]
    \centering\footnotesize
    \includegraphics[width=\linewidth]{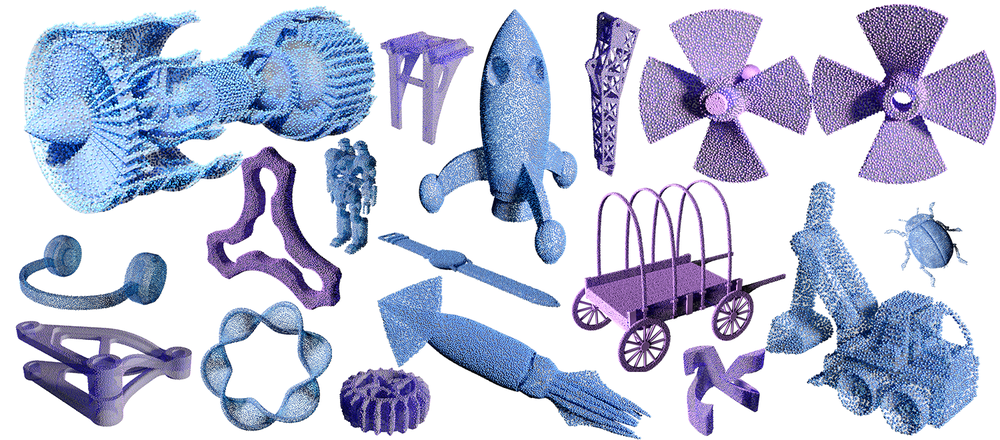}
    \caption{Examples of a few models from our different datasets. We also randomly sample the models with our library.}
    \label{fig:teaser}
\end{figure}

\begin{abstract}
Boundary representation (B-rep) generated from computer-aided design (CAD) is widely used in industry, with several large datasets available~\cite{Koch_2019_CVPR,lambourne2021brepnet,willis2020fusion,willis2021joinable}. However, the data in these datasets is represented in STEP format, requiring a CAD kernel to read and process it. This dramatically limits their scope and usage in large learning pipelines, as it constrains the possibility of deploying them on computing clusters due to the high cost of per-node licenses.

This paper introduces an alternative format based on the open, cross-platform format HDF5 and a corresponding dataset for STEP files, paired with an open-source library to query and process them. Our Python package also provides standard functionalities such as sampling, normals, and curvature to ease integration in existing pipelines. 

To demonstrate the effectiveness of our format, we converted the Fusion 360 dataset and the ABC dataset. We developed four standard use cases (normal estimation, denoising, surface reconstruction, and segmentation) to assess the integrity of the data and its compliance with the original STEP files.

\end{abstract}

\section{Introduction}

Boundary representation (B-rep) is one of the most common formats for representing 3D shapes in solid modeling and computer-aided design, and it is widely used in industry due to its ability to describe precise and complex geometries. B-rep represents shapes as a collection of intersecting parametric surfaces, allowing for the definition of complex smooth surfaces. In recent years, several large datasets have been created containing thousands of B-reps in STEP format~\cite {Koch_2019_CVPR,lambourne2021brepnet,willis2020fusion,willis2021joinable}.

Unfortunately, the data in these datasets is represented in STEP format, requiring a proprietary CAD kernel to read and process it. Additionally, different kernels and different kernel versions are incompatible. This problem led to the flourishing development of CADFix and CADDoctor solutions, whose only purpose is to fix and convert STEP files across different kernels and versions. These barriers affect B-rep usage in large learning pipelines, as they limit the possibility of deploying them on computing clusters due to the formats' highly unstructured and undocumented nature.

This paper introduces an alternative equivalent format, an open-source library to process it, and a corresponding dataset (\Cref{fig:teaser}) for STEP files. Our format is fully specified (\Cref{app:format}) and it is based on the standard half-edge format. To foster cross-language and cross-platform compatibility, we encode it as a dictionary using the HDF5 format; with our format, any application can read and process the data. To ease integration in existing pipelines, we provide a Python package with standard functionalities such as sampling, normals, or curvature. We convert the Fusion 360 and ABC datasets and add another million models from OnShape.

To show the effectiveness of our format and library, we use our library on a series of common learning tasks (i.e, normal estimation, denoising, surface reconstruction, and segmentation), showing how easy it is to use; we confirm that the accuracy obtained by every method is inline with the results reported by the authors. We note that for the classification task in~\citet{fu2023bpnet}, the authors used the triangle meshes in the ABC dataset and used a heuristic to retrieve the parametric information (as it is lost in the meshes). With our library and format, this information is naturally and easily obtainable.

 We hope that our dataset and format will become the new standard benchmark for learning tasks on 3D shapes and that the ability to retrieve parametric information will lead to new, exciting discoveries and progress.

\section{Related Work}

\begin{figure}
    \centering\footnotesize
     \includegraphics[width=0.26\linewidth]{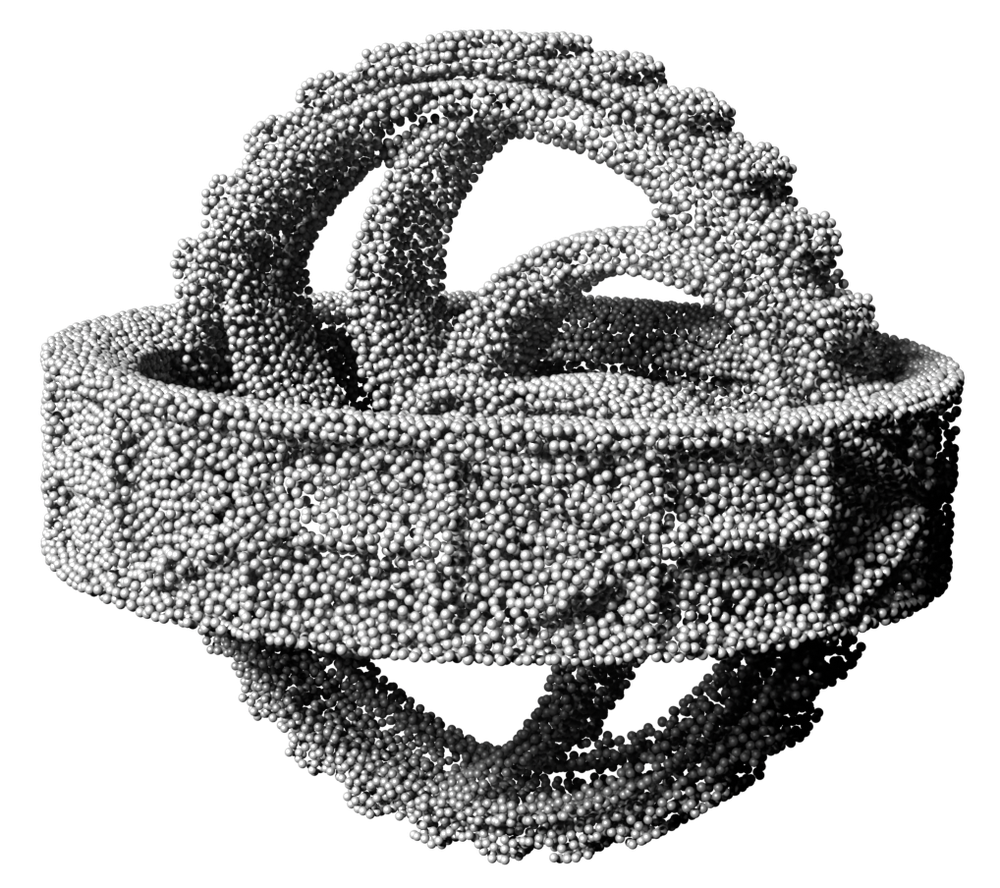}\hfill
     \includegraphics[width=0.26\linewidth]{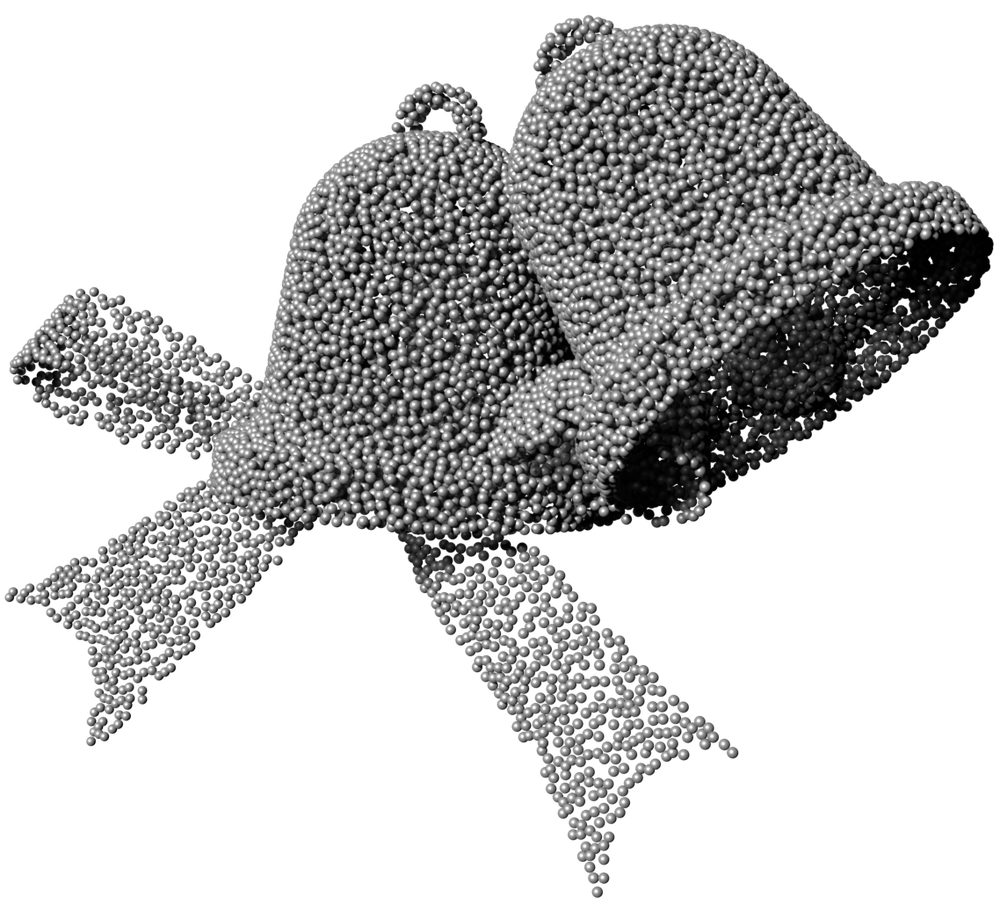}\hfill
     \includegraphics[width=0.17\linewidth]{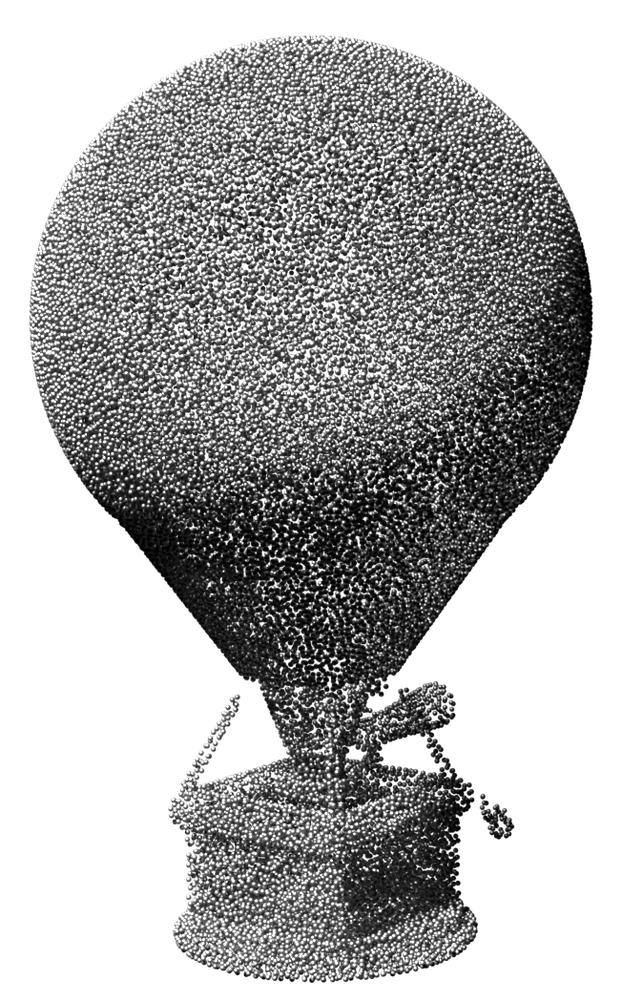} \hfill
     \includegraphics[width=0.21\linewidth]{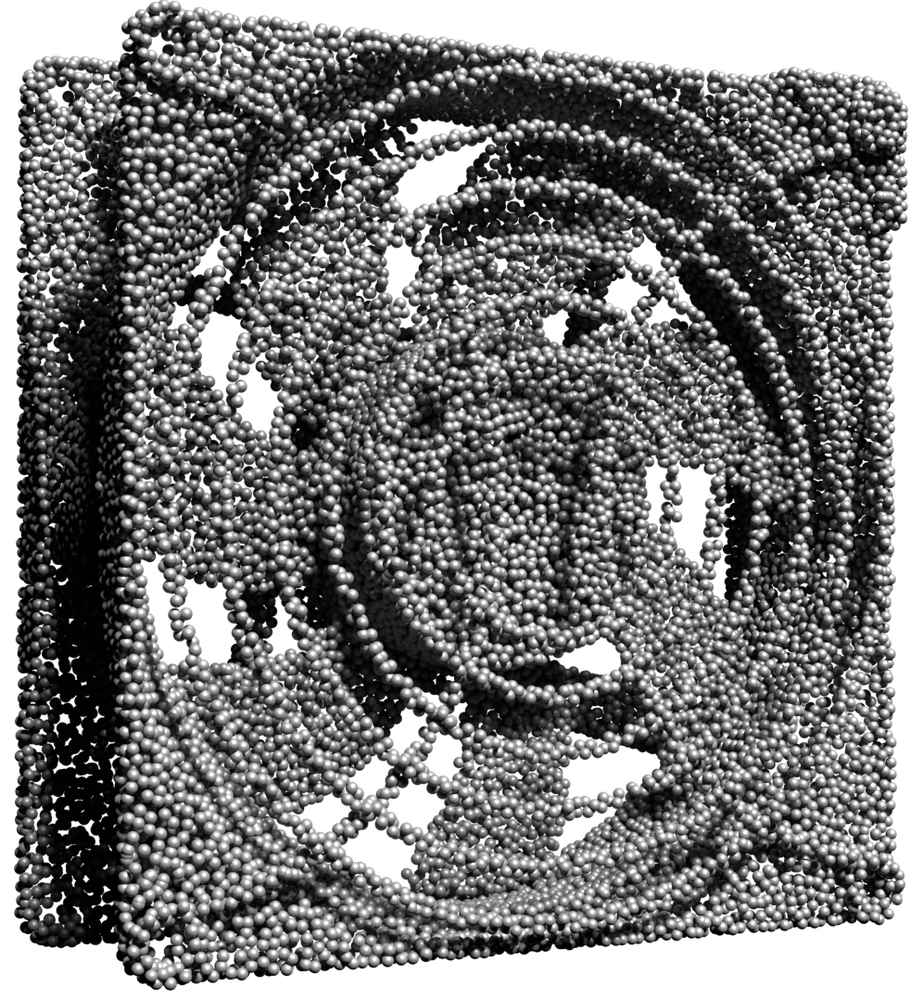}
    \caption{Point-cloud for a model where OpenCascade fails to generate a mesh.}
    \label{fig:mesh-fail}
\end{figure}

Applying machine learning to 3D geometry has created a growing demand for large, richly annotated datasets of 3D shapes in formats that preserve geometric fidelity and support editability. Early shape datasets (\cite{ShapeNet, thingi10, pointnet}) primarily contained annotated meshes or point clouds and were the main drivers of data-driven research on 3D shape understanding and processing. As research in geometric deep learning and computer-aided design (CAD) has progressed, there has been a growing need for representations that go beyond discrete approximations. Recent advances in these fields emphasize the importance of using continuous and smooth B-reps and parametric surfaces~\cite{fu2023bpnet,CADOps,cafl,uvnet}. B-reps are composed of trimmed parametric surfaces and explicitly define the adjacency relationships that connect them into a coherent solid~\cite{lambourne2021brepnet}. They offer the advantage of richer semantic annotations (e.g., the type and shape of components, and how they are assembled), enabling learning tasks that incorporate both geometry and the generative design process.

Unlike mesh-based representations, which discretize geometry and may lose important structural information, B-reps retain the exact geometry and topology defined during CAD modeling~\cite{willis2021joinable}. This feature enables precise querying and captures high-level design semantics, making B-reps well suited for applications such as reconstruction~\cite{Brep2Seq} and constraint inference~\cite{cafl}. Furthermore, since multiple CAD models can result in identical sampled meshes, mesh-based data tends to be more ambiguous. In contrast, B-reps provide a more reliable foundation for tasks requiring interpretability and reversibility~\cite{CADOps,DeepCAD}.

Recognizing these advantages, recent efforts have focused on building datasets natively supporting B-reps or CAD formats such as STEP. ABC~\cite{Koch_2019_CVPR} was among the first to collect one million 3D STEP files. This dataset sparked growing interest in developing more datasets that support native B-reps~\cite{lambourne2021brepnet, uvnet}, enabling the design of neural architectures that operate directly on these structures rather than on their triangulated approximations. Subsequently, the Fusion 360 Gallery~\cite{willis2020fusion,willis2021joinable} dataset introduced thousands of STEP file sequences, along with the sequence operations used to construct the final model.

As a result, this new perspective has inspired the creation of additional datasets and benchmarks, such as DeepCAD~\cite{DeepCAD}, which provides over 170,000 models with construction sequences. Brep2Seq~\cite{Brep2Seq} introduces a large-scale collection of auto-synthesized, feature-based CAD models. More recently, datasets such as Param20K~\cite{cafl} have enriched parametric data with explicit annotations.

\section{Library}

We developed two Python libraries: one for converting, \textsc{steptohdf5}, and another for processing, \textsc{ABS}, the datasets. Both libraries use the HDF5 format and NumPy for data encoding and are available via \texttt{pip}\footnote{\url{https://github.com/better-step/abs}}.

\subsection{Steptohdf5}
\textsc{steptohdf5} uses OpenCascade~\cite{opencascade} to parse and extract the geometric and topological information from the STEP file and convert it into our dictionary-based format (\Cref{app:format}). In our format, we decompose every file into a sequence of parts; every part contains geometry, topology, and a mesh. We note that the meshing algorithm in OpenCascade is not fully robust, and approximately 5\% of the models fail to produce a mesh (\Cref{fig:mesh-fail}). The geometry includes all parametric representations, such as curves and surface patches, while the topology defines the connectivity between these geometric entities, specifying how edges, faces, and shells are assembled into a coherent structure. 

\paragraph{Geometry.}
Geometry contains the geometric information in the form of a list of two- and three-dimensional curves and surfaces and a matrix of vertices. Each geometry entity has its own parametric domain (a line for curves and a rectangle for surfaces) and the parameters that define it. For instance, a B-spline curve has a set of control points and knots, while a plane has two axes and a location.

\begin{figure}
    \centering
    \includegraphics[width=\linewidth]{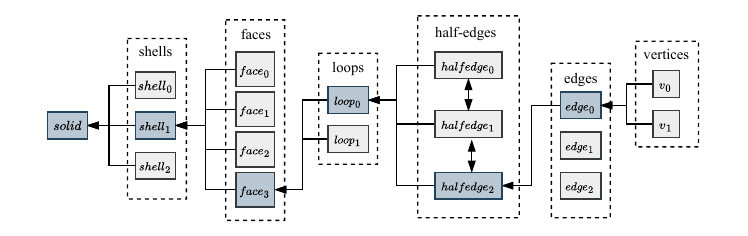}
    \caption{Hierarchical structure of our format. From the root structure (solid) to the leaf (vertices).}
    \label{fig:tree}
\end{figure}

\paragraph{Topology.}
The topology contains a hierarchical structure of the STEP file (\Cref{fig:tree}). To save space and maintain consistency, we store only the top-down relationships (e.g., a face contains loops, but a loop does not store the face it belongs to) and use our library to recover the reverse links when needed. The root of the tree contains the solid; it contains one unique field to store the list of shells. Each shell contains a list of the faces in that shell and an orientation flag. In the case of a manifold solid, the orientation flag will always be true. In the case of a non-manifold cell complex, where multiple shells may share one face, the flag represents whether the face normal needs to be flipped to point outwards from the solid volume the shell bounds. Both solid and shell are purely topological entities and do not have a geometric counterpart (inset).
\begin{wrapfigure}{l}{0.25\linewidth}
    \centering
    \includegraphics[width=\linewidth]{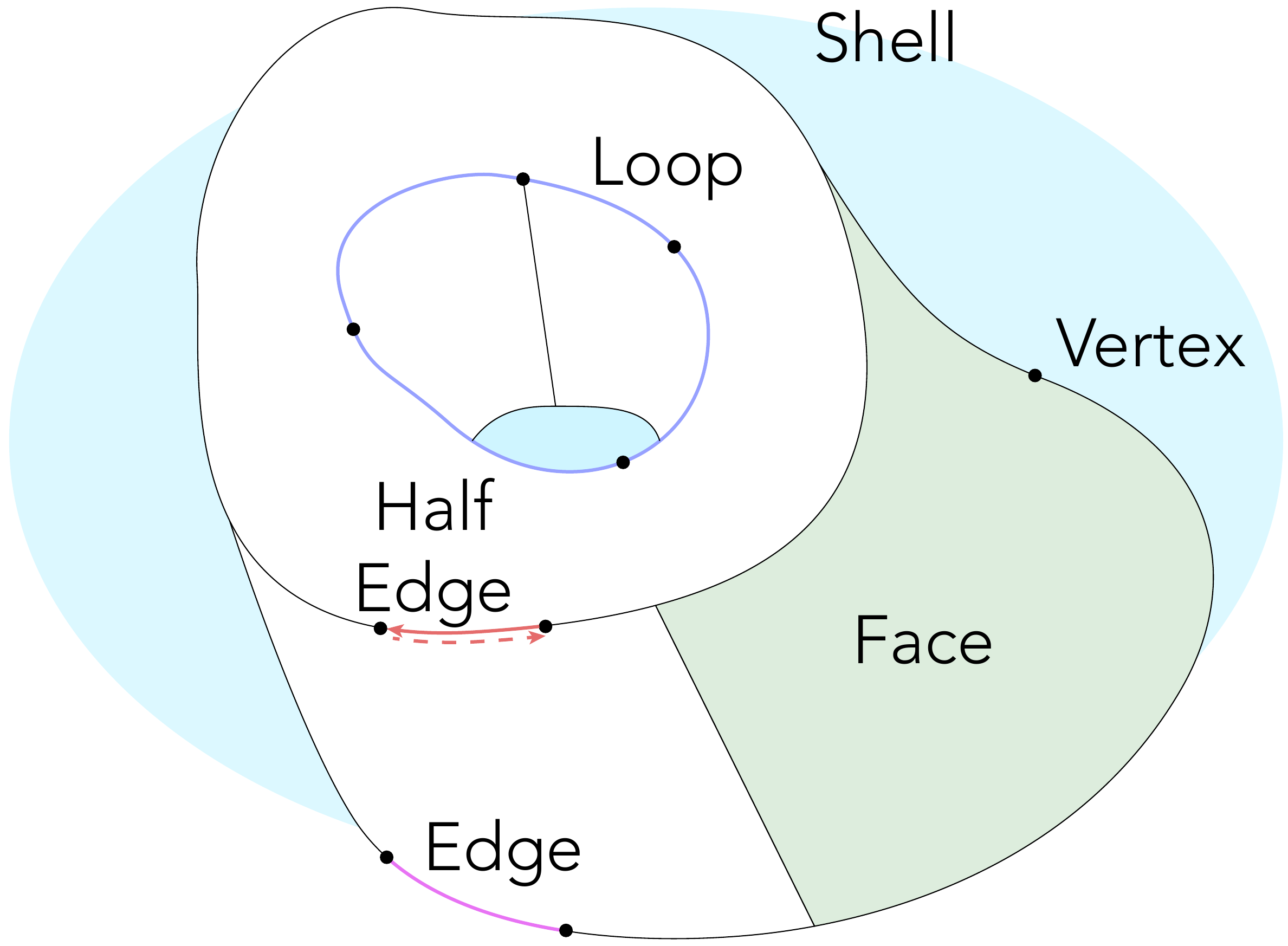}
    \vspace{-3em}
\end{wrapfigure}

A face represents a patch and contains a surface index pointing to the corresponding surface in the geometry. Additionally, it includes the orientation and a list of loops. Each loop is a closed poly curve representing the trimming of the face. It consists of a list of half-hedges that can be shared by two edges. A half-edge also includes the index of a two-dimensional curve in the geometry file, its mates (the opposite half-edge), an orientation flag, and the associated edge. As the edge contains the pointer to the three-dimensional curve, the orientation flag is used to properly orient the half-edge in the loop.

\subsection{ABS}
\begin{figure}
    \centering\footnotesize
    \parbox{.8\linewidth}{\centering
    \parbox{.24\linewidth}{\includegraphics[width=\linewidth]{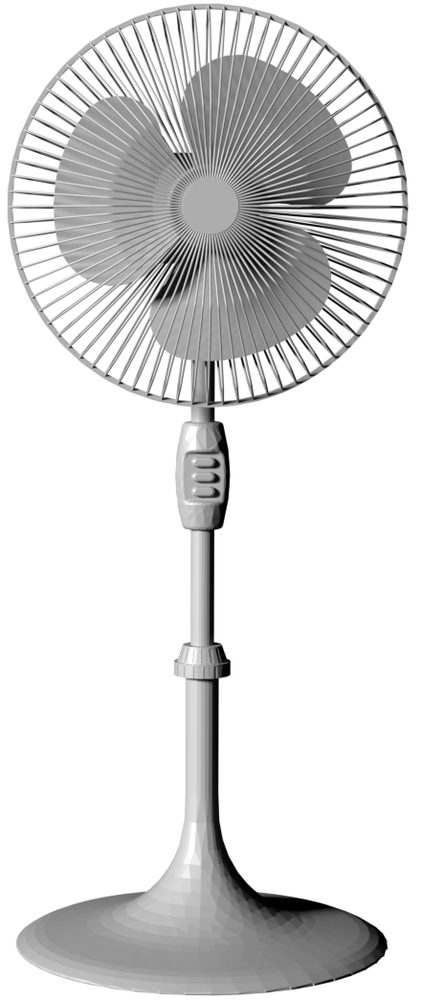}}\hfill
    \parbox{.75\linewidth}{\centering
    \includegraphics[width=0.32\linewidth]{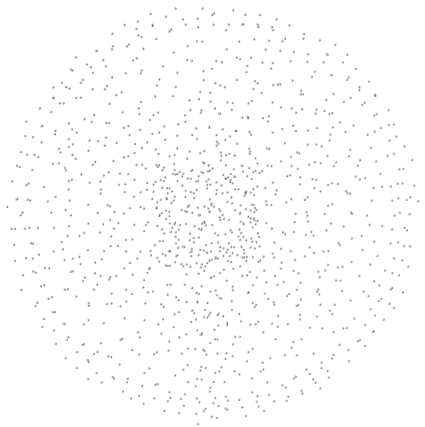}\hfill
    \includegraphics[width=0.32\linewidth]{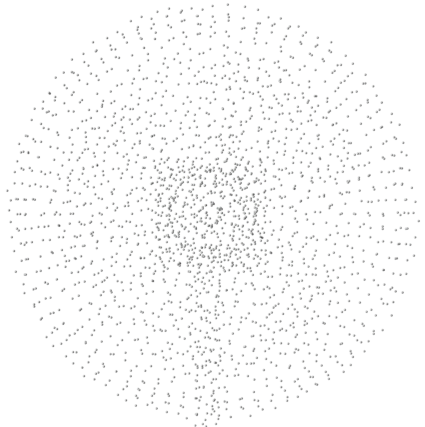}\hfill
    \includegraphics[width=0.32\linewidth]{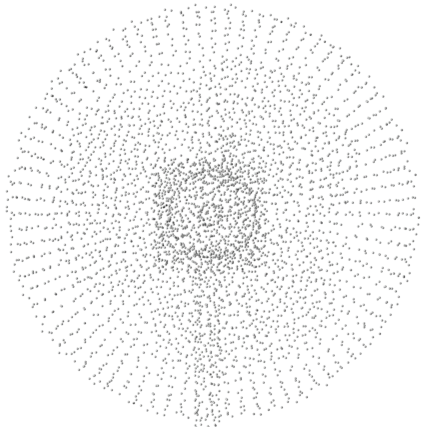}\par
    \parbox{.32\linewidth}{\centering 2\,000 points.}\hfill
    \parbox{.32\linewidth}{\centering 4\,000 points.}\hfill
    \parbox{.32\linewidth}{\centering 8\,000 points.}\\[1em]
    \includegraphics[width=0.32\linewidth]{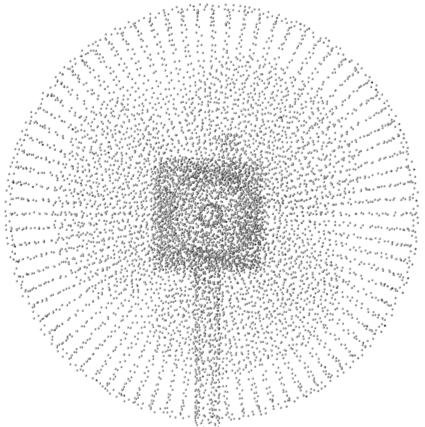}\hfill
    \includegraphics[width=0.32\linewidth]{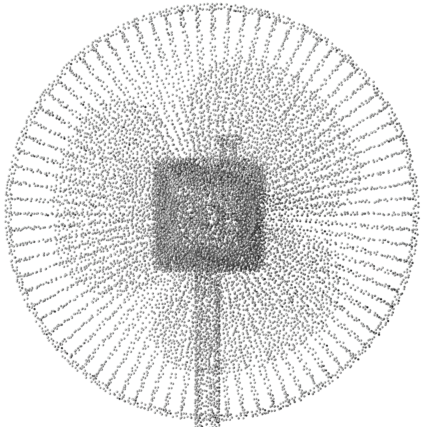}\hfill
    \includegraphics[width=0.32\linewidth]{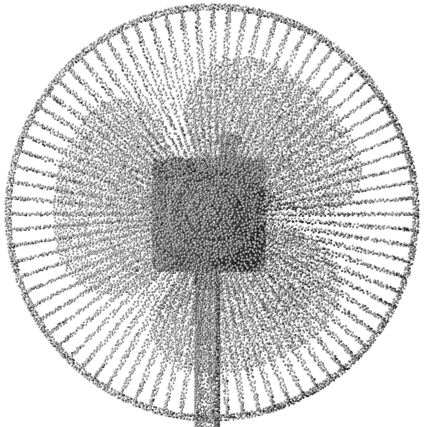}
    \parbox{.32\linewidth}{\centering 16\,000 points.}\hfill
    \parbox{.32\linewidth}{\centering 32\,000 points.}\hfill
    \parbox{.32\linewidth}{\centering 64\,000 points.}
    }}
    \caption{Example of a complex fan model (left) sampled with an increasing number of points. As the resolution increases, the small details become visible.}
    \label{fig:sampling-res}
\end{figure}

\begin{figure}
    \centering\footnotesize
    \includegraphics[width=0.26\linewidth]{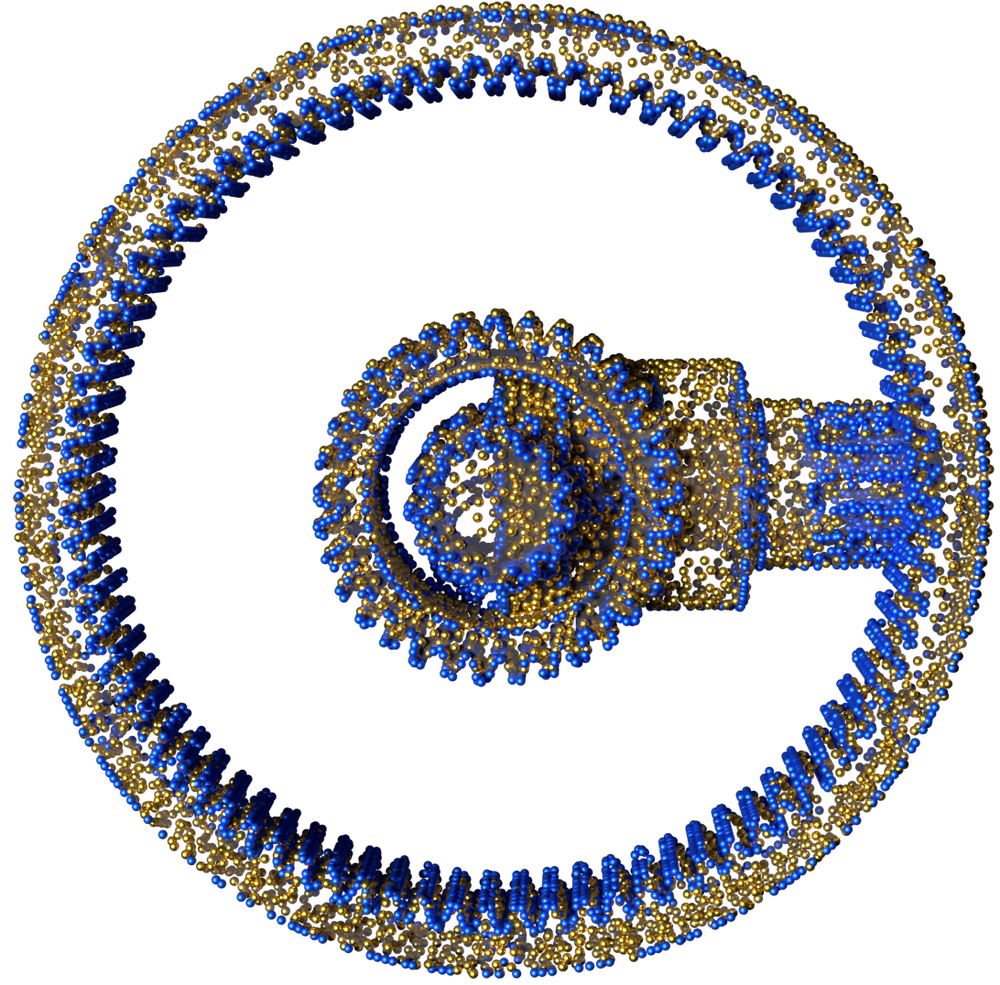}\hfill
    \includegraphics[width=0.26\linewidth]{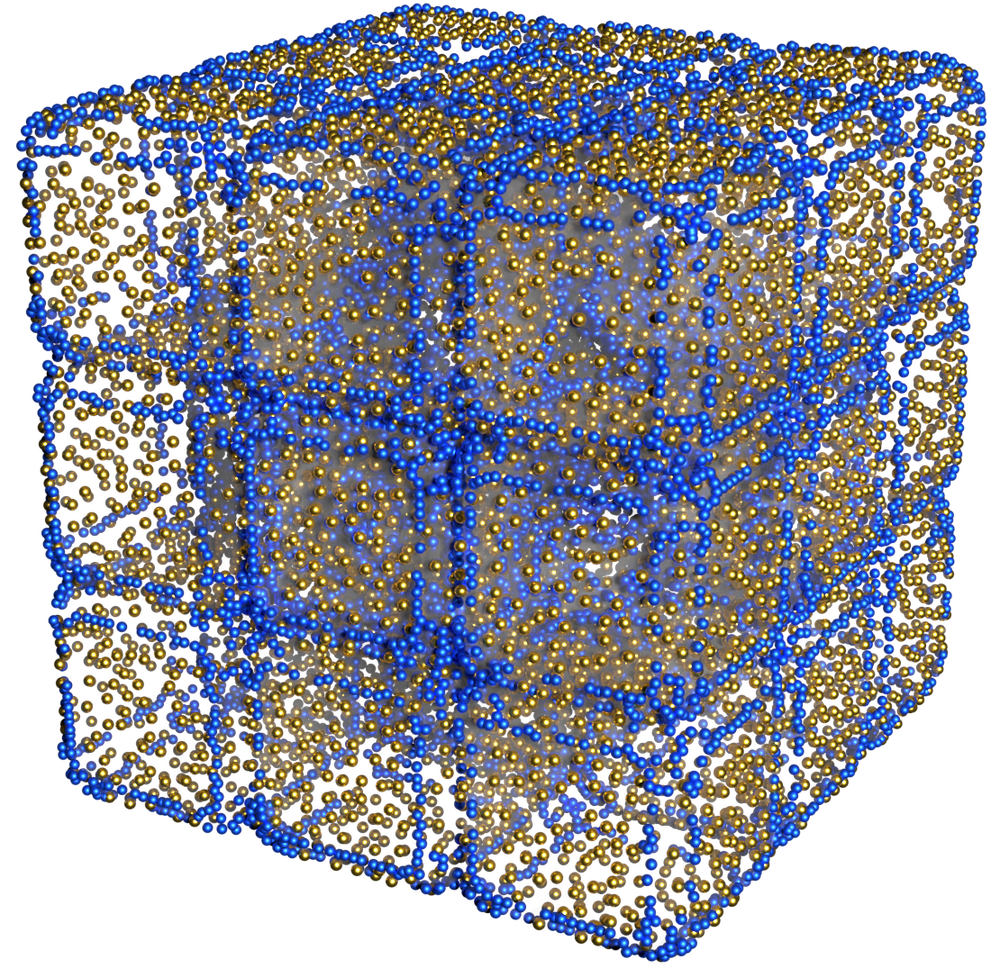}\hfill
    \includegraphics[width=0.18\linewidth]{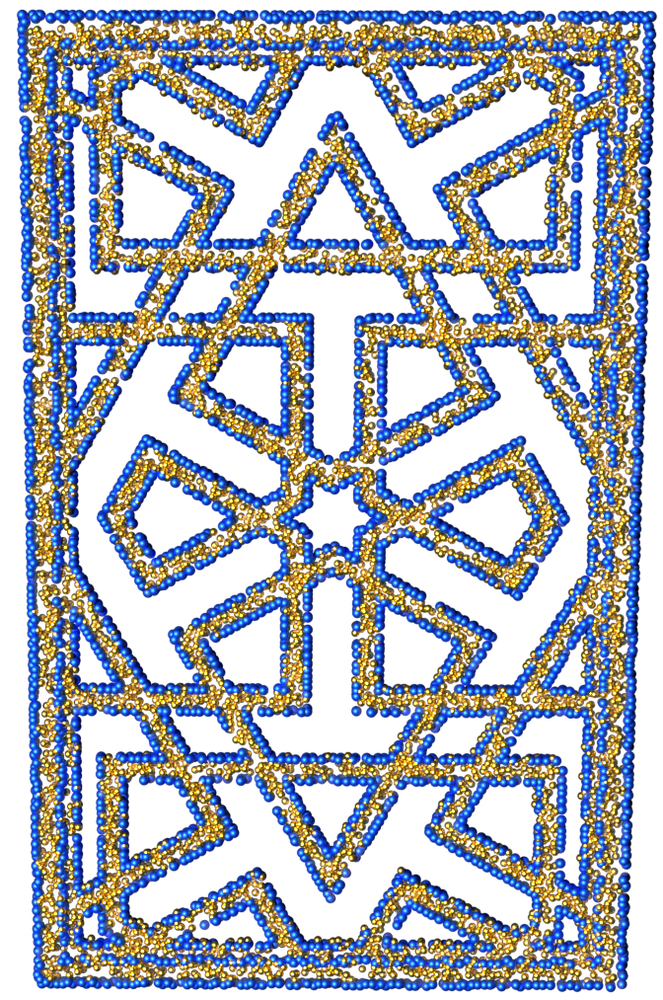} \hfill
    \includegraphics[width=0.26\linewidth]{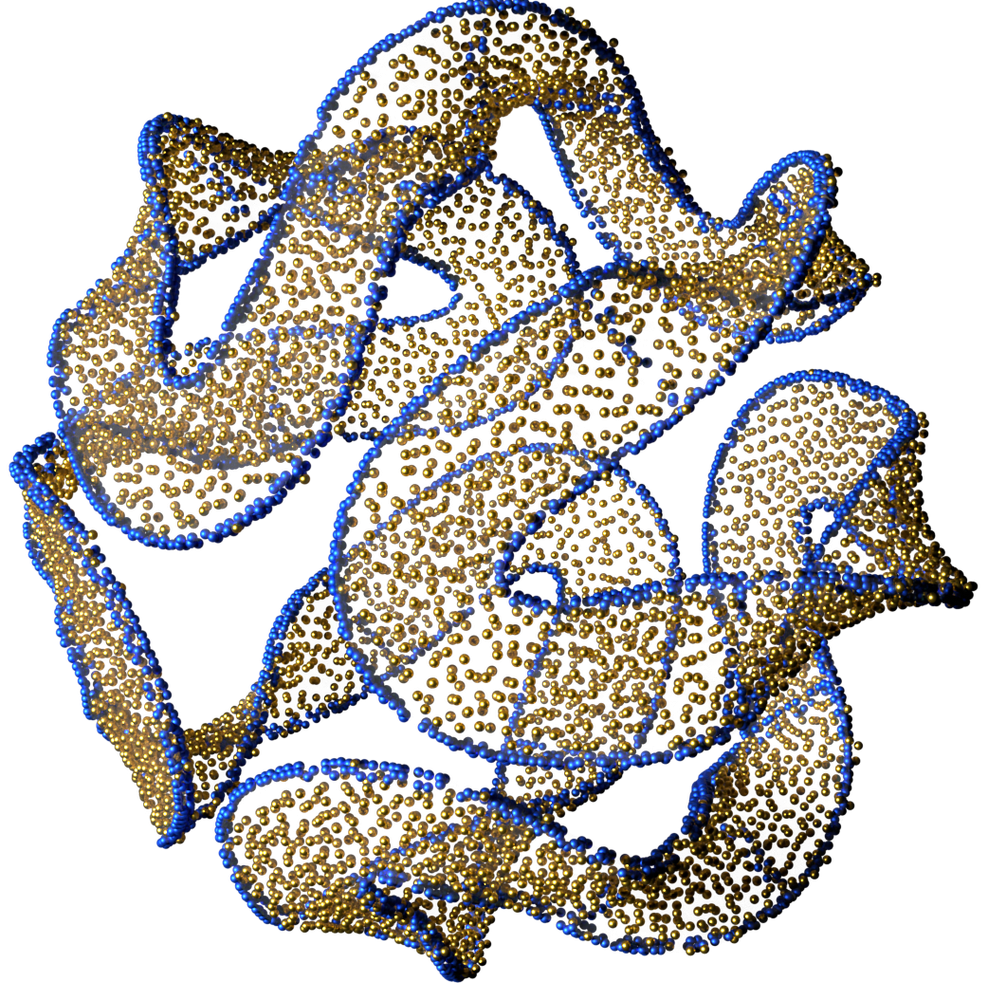}
    \caption{Example of point clouds sampled using Listing~\ref{lst:sample}; we highlight the feature edges in yellow.}
    \label{fig:fedge}
\end{figure}

Our format only contains equivalent information to the B-rep data, and using it directly in an application might be challenging. We developed a library that allows processing, navigating, and extracting features from the dataset to facilitate its usage. 
\textsc{ABS} allows reading and navigating the HDF5 files as a standard half-edge data structure and can generate random points sampled directly from the \emph{continuous parametric} shapes and evaluate parametric derivatives (\Cref{fig:sampling-res}).
We provide a simple \texttt{read\_parts} and \texttt{read\_meshes} to read the parts and meshes respectively from an 
input HDF5 file and a function \texttt{sample\_parts}
that uses a lambda function to decide which information to extract (Section~\ref{sec:cases} shows more concrete examples). 
The lambda function has access to the current part, the current topological entity (either a face or an edge), and the random points in the parametric domain. Its responsibility is to return data associated with the points (e.g., a normal or label), or \texttt{None} if the entity must be skipped. Listing~\ref{lst:sample} shows a typical example of how to use  \textsc{ABS}, we use \texttt{read\_parts} to read the file and \texttt{compute\_labels} and \texttt{sample\_parts} to sample the shape and obtain a binary label to mark feature edges or patches (\Cref{fig:fedge}). 

\begin{lstlisting}[language=Python, caption=Example of computing normal at every point., frame=lines, label=lst:sample]
from abs import read_parts, sample_parts


def compute_labels(part, topo, points):
    if topo.is_face(): return 1
    else: return 0

parts = read_parts(file_path)
P, S = sample_parts(parts, num_samples, compute_labels)
\end{lstlisting}

Since the mesh is not connected with the topology and the geometry of the B-rep, we provide a utility function \texttt{read\_meshes} to retrieve the meshes as point-triangle dictionary, one per part per face (Listing~\ref{lst:meshes}). Note that, as not every face can be meshed, the pair can be \texttt{None}.
For instance, \texttt{meshes[0][9]} contains the mesh of the 10th face of the first part. Additionally, we have a simple function that concatenates every meshed patch into a unique, consistent mesh.

\begin{lstlisting}[language=Python,frame=lines,  caption=Example of extracting the mesh from a file., label=lst:meshes]
from abs.utils import read_meshes, get_mesh

meshes = read_meshes(file_path)
V, F = get_mesh(meshes)
\end{lstlisting}

\section{Dataset}

\begin{figure}
    \centering\footnotesize
     \includegraphics[width=0.16\linewidth]{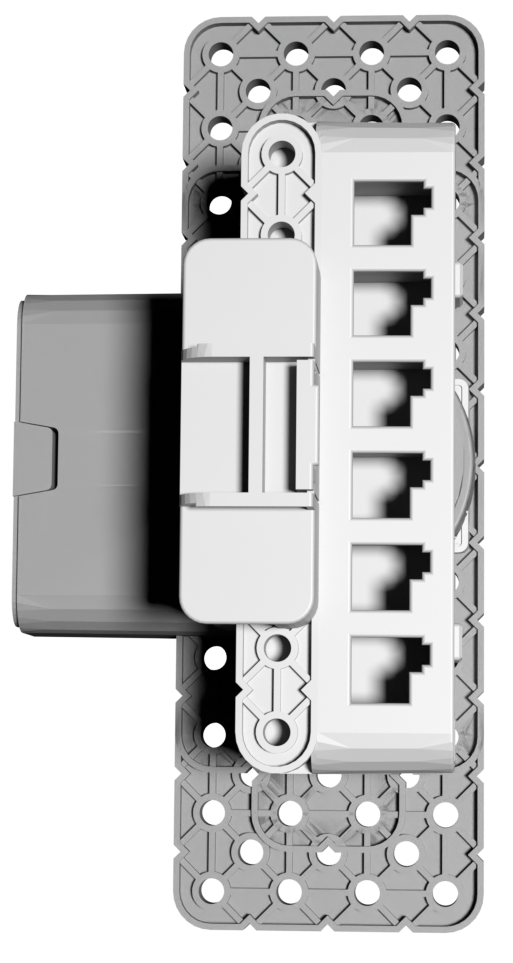}\hfill
     \includegraphics[width=0.29\linewidth]{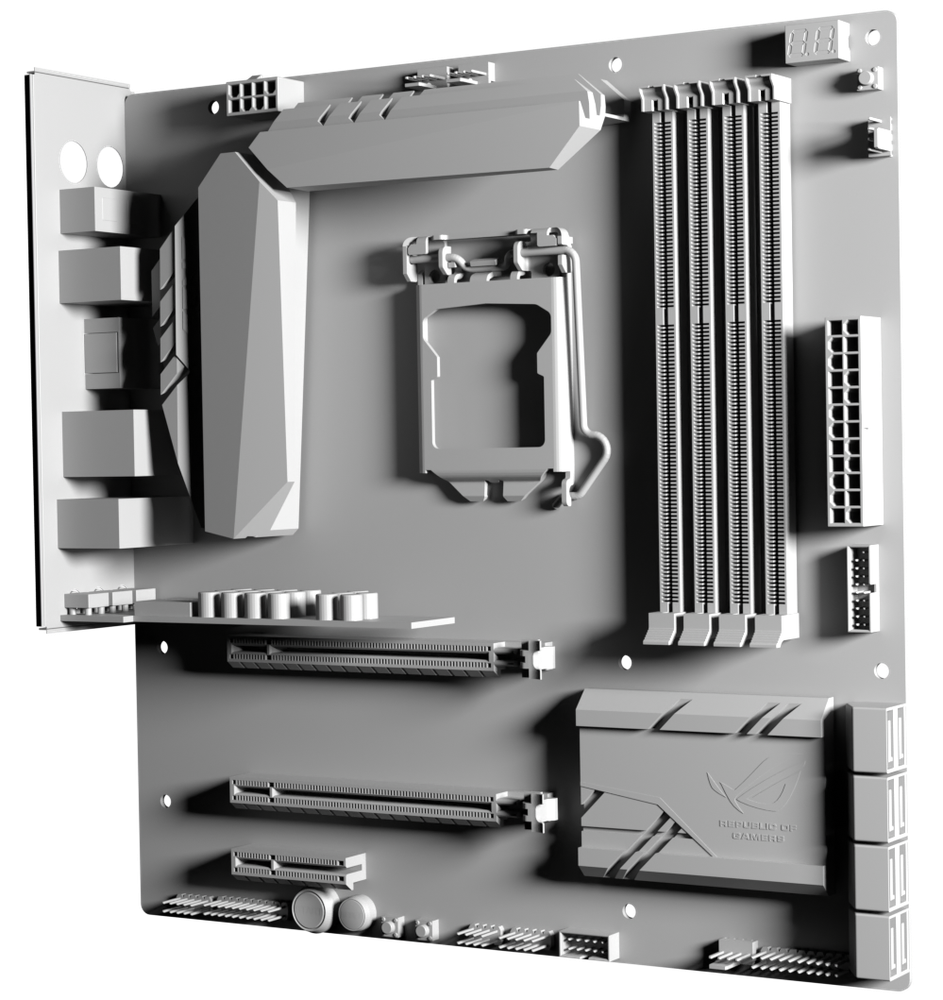}\hfill
     \includegraphics[width=0.24\linewidth]{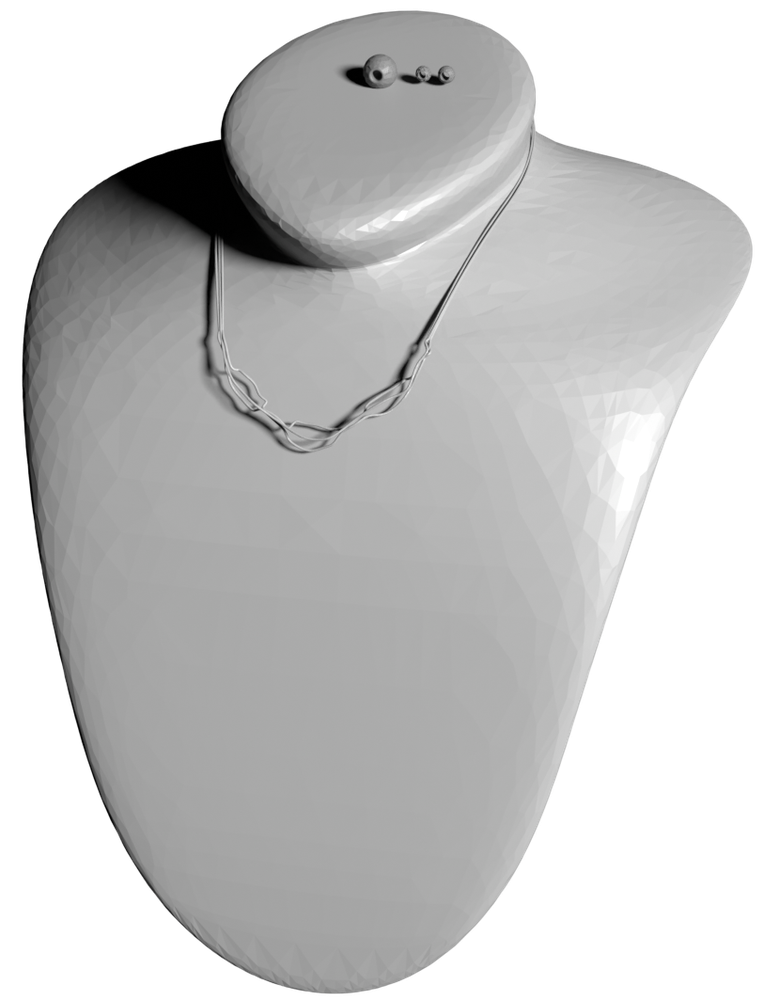}\hfill
     \includegraphics[width=0.29\linewidth]{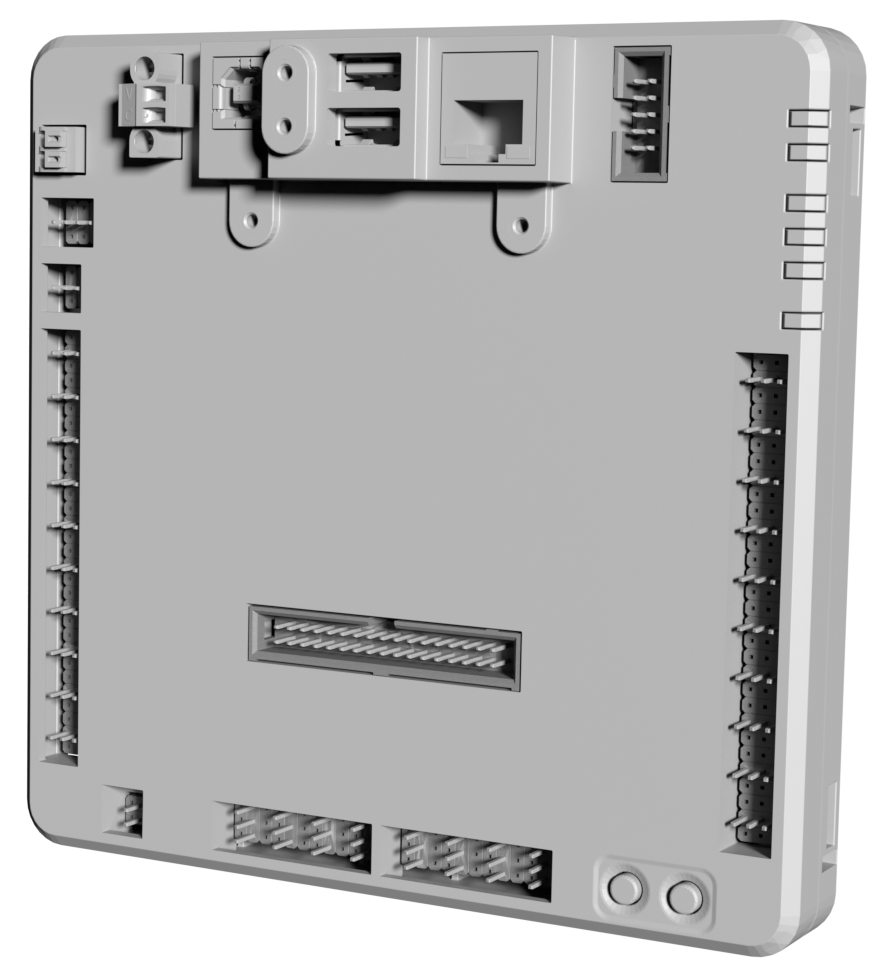}
    \caption{Example of models with thousands of patches.}
    \label{fig:large-models}
\end{figure}

\begin{figure}
    \centering\footnotesize
    \includegraphics[width=\linewidth]{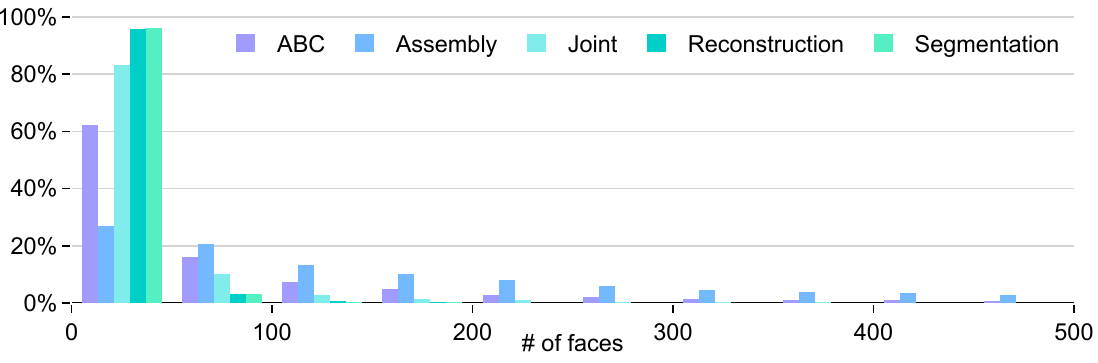}
    \caption{Distribution of of faces per model for the different datasets.}
    \label{fig:num_patches}
\end{figure}

\begin{figure}
    \centering\footnotesize
    \includegraphics[width=\linewidth]{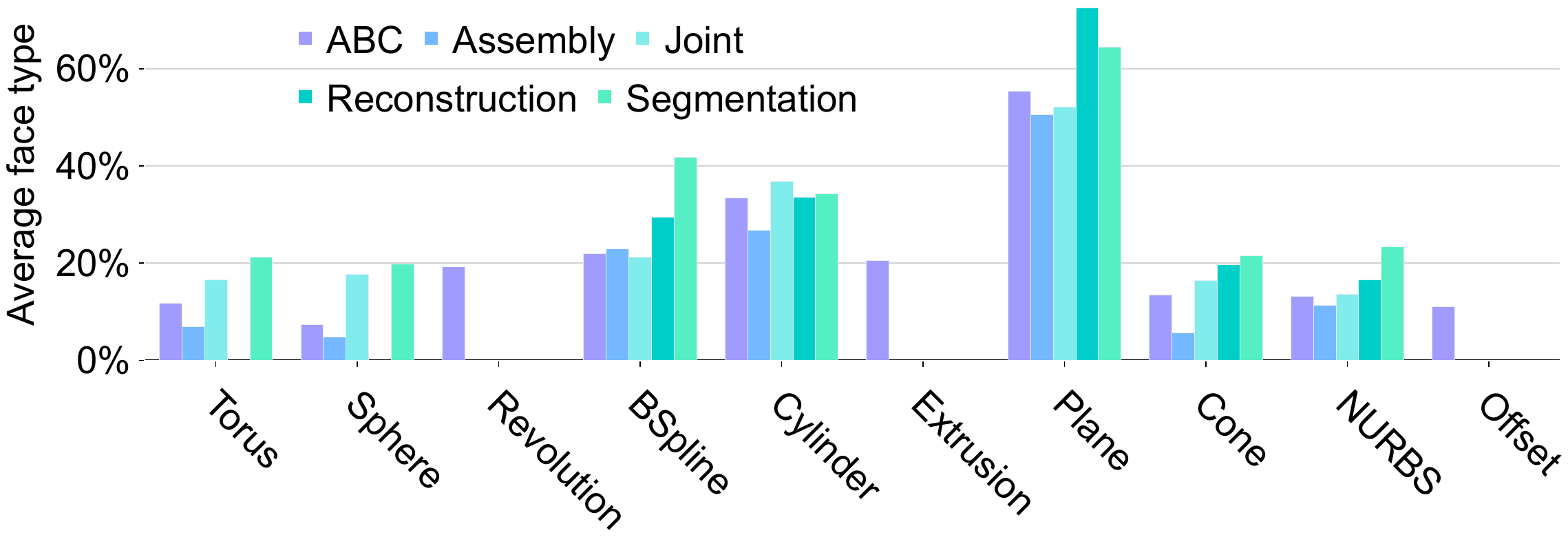}
    \includegraphics[width=\linewidth]{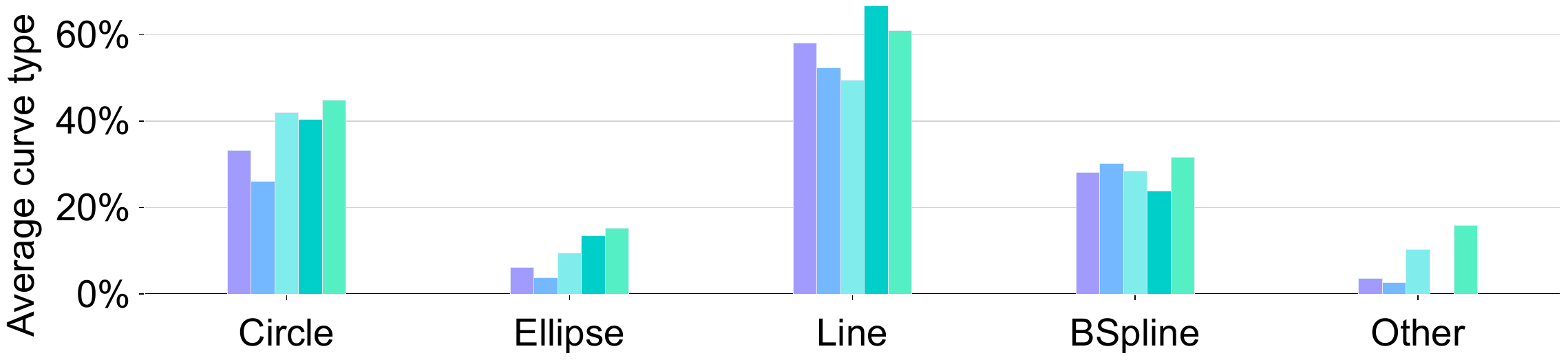}
    \caption{Average distribution of faces (top) and curves (bottom) types across the different datasets.}
    \label{fig:type}
\end{figure}

Our dataset\footnote{\url{https://www.frdr-dfdr.ca/repo/dataset/d54b95e0-bc14-4236-b50b-922e5bf4ba7d}}  includes one million models from ABC \cite{Koch_2019_CVPR}, as well as the Assembly (8\,251 models, 16\,2707 parts), Joint (23\,029 parts), Reconstruction (27\,958 parts), and Segmentation (35\,680 parts) from Fusion 360 dataset~\cite{willis2020fusion,willis2021joinable}. We converted the data on cluster nodes equipped with Intel E5 v4 Broadwell @ 2.2GHz CPUs. On average, converting a single model takes a few seconds, and processing the entire dataset requires approximately one CPU year. We computed statistics on a random selection of 4\,000 models for ABC, on the assembled models for Assembly, and on the entire dataset for Joints, Reconstruction, and Segmentation.

On average, the models contain 137 patches (ABC: 236, Assembly: 590, Joint: 37, Reconstruction: 15, and Segmentation: 15) with models with more than 30\,000 patches (\Cref{fig:large-models}). The different datasets contain models of varying sizes (\Cref{fig:num_patches}); Assembly is the largest overall (even though ABC includes the model with the most faces), while Segmentation is the smallest. All models in the dataset contain about 50\% planes; with the Reconstruction dataset having the highest proportion at 72\% (\Cref{fig:type}, top). Only the ABC dataset includes a small number of offset, revolution, and extrusion surfaces. Similarly, most models consist primarily of lines and circles, while the Segmentation dataset includes the highest number of unrecognized curves marked as "other" (\Cref{fig:type}, bottom).

The meshing algorithm in OpenCascade is not robust and occasionally fails (i.e., some of the patches have no mesh), with a failure rate of 1.56\% for the ABC dataset,  8.82\% for the Assembly dataset, 
1.04\% for the Joint dataset, 0.02\% for the Reconstruction dataset, and 0.07\% for the Segmentation dataset.

\section{Use Cases}\label{sec:cases}

We showcase the simplicity and versatility of our library by generating data for four point-cloud-based machine learning tasks: normal estimation, denoising, reconstruction, and segmentation. For all use cases, we use the same code as in Listings~\ref{lst:sample} except that we write a task-specific lambda function. Note that we do not fine-tune or retrain the models; we evaluate them directly using our dataset.

\paragraph{Normal estimation.}A classical learning problem involves estimating normals from a point cloud, which requires a dataset of point clouds paired with ground truth normals. This can be easily computed from our dataset using the function in Listings~\ref{algo:normals}. We evaluate the DeepFit model~\cite{ben2020deepfit} on 8,000 points generated with \textsc{ABS}, sampled from 200 randomly selected models in the ABC dataset. Although the model was trained on piecewise linear geometries (i.e., meshes), it performs well in estimating smooth normals. The percentage of good points (PGP), ignoring normal orientation, is 65.73\%, 79.45\%, and 90.28\% for angular thresholds of $5^\circ$, $10^\circ$, and $30^\circ$, respectively.

\begin{lstlisting}[language=Python,frame=lines,  caption=Extracting the normals., label=algo:normals]
def compute_normals(part, topo, points):
    if topo.is_face(): return topo.normal(points)
    else: return None # No normals for edges
\end{lstlisting}

\begin{figure}
    \centering\footnotesize
    \includegraphics[width=0.33\linewidth]{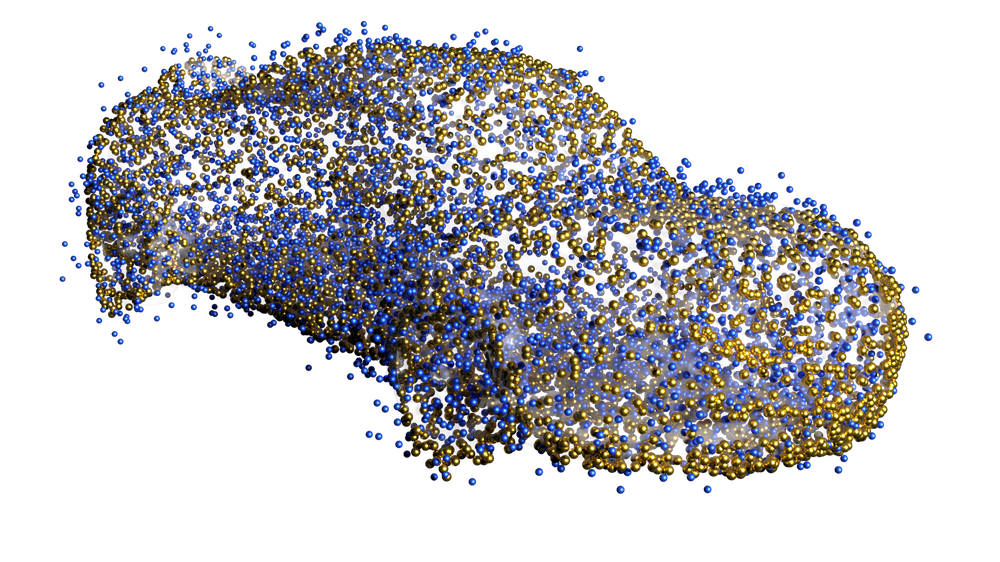}\hfill
    \includegraphics[width=0.33\linewidth]{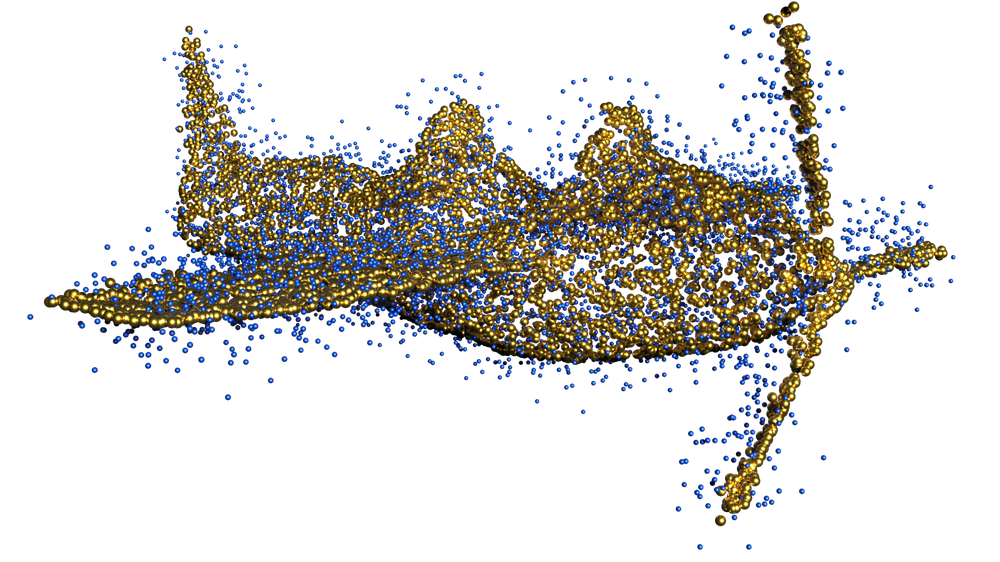}\hfill
    \includegraphics[width=0.33\linewidth]{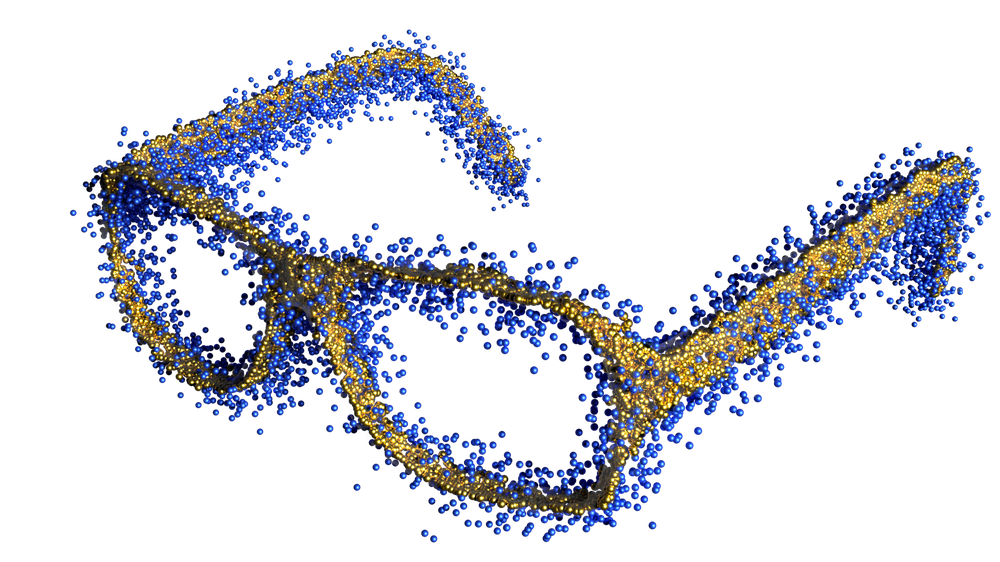}
    \caption{Example of denoising point cloud using PathNet~\cite{wei2024pathnet}.}
    \label{fig:denoise}
\end{figure}

\paragraph{Denoising.}

We use the recently published PathNet~\cite{wei2024pathnet} to denoise point clouds. The model consists of a two-stage deep and reinforcement learning pipeline. It dynamically selects the optimal denoising path for each point using a reinforcement learning-based routing agent that adapts to local noise levels and geometric complexity. The model requires only the input noisy point cloud and outputs the denoised result for both training and evaluation (\Cref{fig:denoise}). Ground truth can be generated using Listings~\ref{algo:denoise}, with noise added afterward. While~\citet{wei2024pathnet} train their model using mesh-sampled data, our dataset and library allow direct sampling from smooth parametric surfaces. Despite this slight difference, the method performs comparably when applied to our dataset. We selected 1\,000 models from the Assembly dataset, sampled each with 6\,000 random points, and added varying levels of Gaussian noise. To evaluate performance, we sampled each model with 10\,000 points and computed the MSE (in units of $\times 10^{-3}$) at different noise levels: 33.9, 34.07, and 35.79 for noise levels of 0.5\%, 1\%, and 1.5\%, respectively.

\begin{lstlisting}[language=Python,frame=lines,  caption=Extracting just the points., label=algo:denoise]
def get_points(part, topo, points):
    if topo.is_face(): return 1 # Return dummy value
    else: return None 
\end{lstlisting}

\begin{figure}
    \centering\footnotesize
    \includegraphics[width=0.18\linewidth]{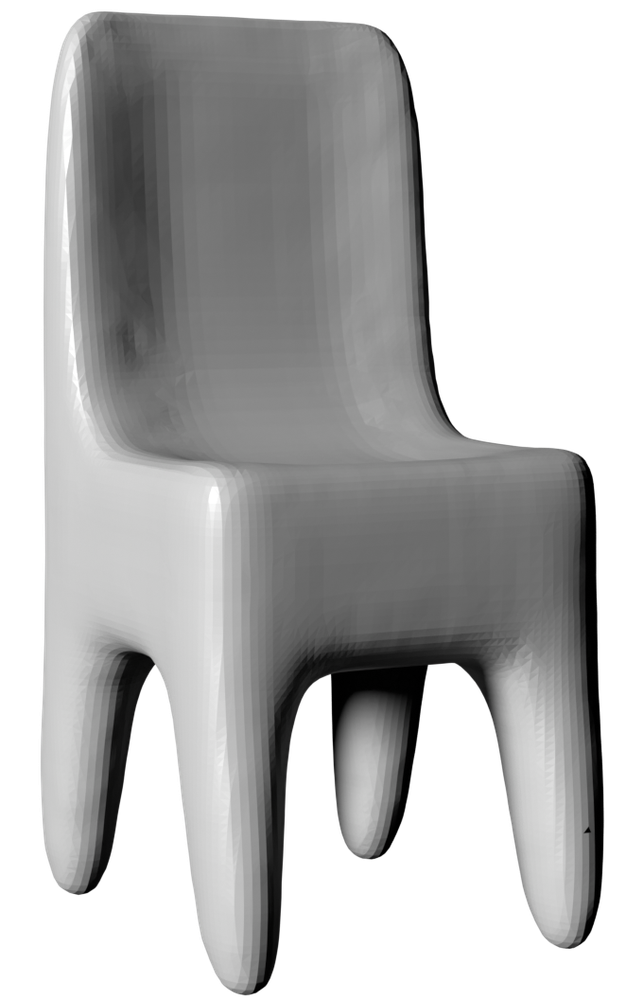}\hfill
    \includegraphics[width=0.18\linewidth]{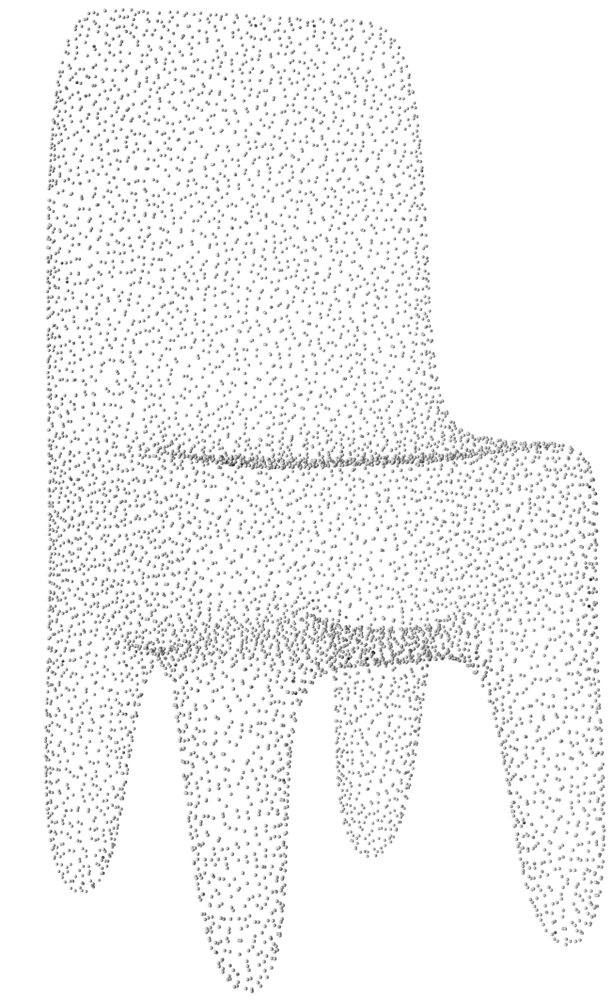}\hfill
    \includegraphics[width=0.16\linewidth]{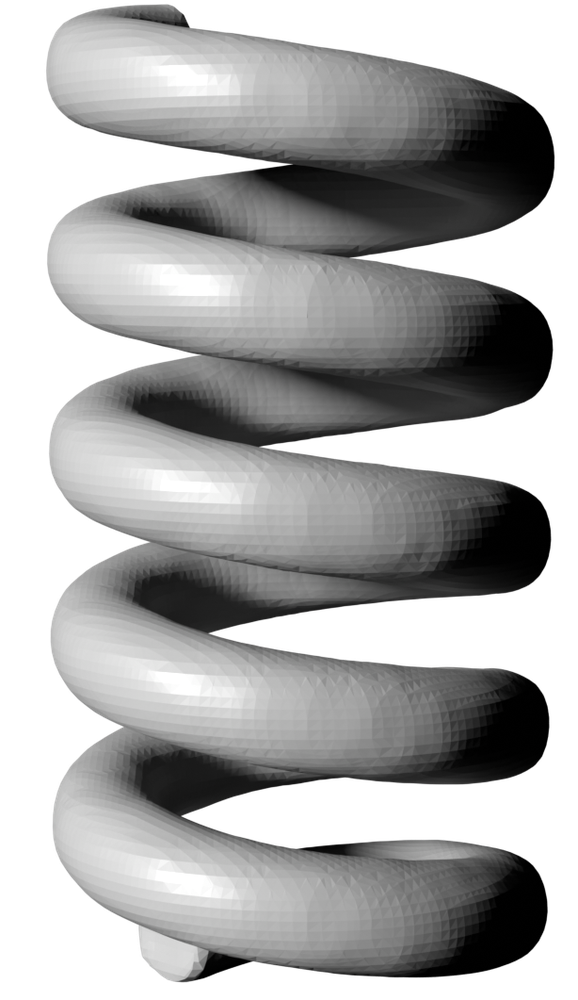} \hfill
    \includegraphics[width=0.16\linewidth]{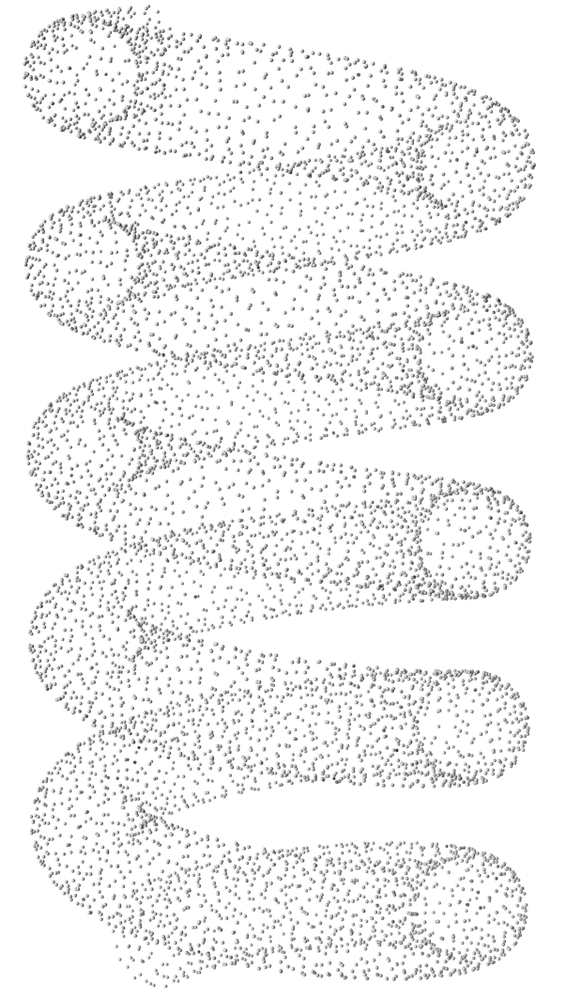}
    \caption{Example of reconstructed surfaces from randomly sampling our dataset using NKSR~\cite{huang2023neural}.}
    \label{fig:reconstuction}
\end{figure}

\paragraph{Surface Reconstruction.}
We selected Neural Kernel Surface Reconstruction (NKSR)~\cite{huang2023neural} as an example method for reconstructing meshes from a potentially noisy point clouds (\Cref{fig:reconstuction}). This approach represents surfaces as a zero-level set of a neural kernel field fitted to oriented point clouds via a gradient-based energy formulation, using only points and corresponding normals for supervision for training. Both training and evaluation datasets can be generated with the same code as in Listings~\ref{algo:normals}, using denser sampling for training. For our experiments, we selected 1\,000 models from the Assembly dataset, sampled these models with varying numbers of points, and added random Gaussian noise. To evaluate reconstruction quality, we computed Chamfer distance and F-Score metrics between the reconstructed surfaces and a dense sampling (15\,000 points) of the parametric surfaces. Our findings, summarized in \Cref{tab:reconstruction}, indicate slightly lower reconstruction quality compared to results reported by \citet{huang2023neural}. This reduction in quality is likely due to our direct sampling of parametric surfaces instead of using pre-existing meshes.

\begin{table}
    \centering
    \scriptsize  %
    \caption{Evaluation metrics for surface reconstruction with varying sample sizes and noise levels ($\sigma$). Chamfer distance ($d_C$) values are scaled by $10^3$.}
    \label{tab:reconstruction}
    \begin{tabular}{lccc|ccc|ccc}
        \toprule
        & \multicolumn{3}{c|}{\textbf{4000 samples}} 
        & \multicolumn{3}{c|}{\textbf{6000 samples}} 
        & \multicolumn{3}{c}{\textbf{8000 samples}} \\
        \cmidrule(r){2-4} \cmidrule(r){5-7} \cmidrule(r){8-10}
        & $\sigma=0$ & $\sigma=0.005$ & $\sigma=0.025$ 
        & $\sigma=0$ & $\sigma=0.005$ & $\sigma=0.025$ 
        & $\sigma=0$ & $\sigma=0.005$ & $\sigma=0.025$ \\
        \midrule
        $d_{c}$   & 3.39 & 1.91 & 2.02 & 3.42 & 1.48 & 3.75 & 1.63 & 2.20 & 2.56 \\
        F-Score   & 84.62 & 85.61 & 60.0 & 87.02 & 84.56 & 61.82 & 85.81 & 85.59 & 63.0 \\
        Precision & 82.09 & 82.10 & 52.65 & 83.78 & 79.56 & 54.20 & 81.11 & 80.79 & 55.01 \\
        Recall    & 91.13 & 92.23 & 72.85 & 93.34 & 94.31 & 75.17 & 95.16 & 95.16 & 77.25 \\
        \bottomrule
    \end{tabular}
\end{table}

\paragraph{Segmentation.}
A complex problem consists of correctly labelling points in a point cloud based on the geometric primitive. For instance, it automatically detects which points belong to a plane or a cylinder. We can use our library to compute the labels as we sample the surface, using a different callback that converts the surface type into the label (Listings~\ref{lst:segmentation}). We use the BPNet~\cite{fu2023bpnet} model that uses labelled points as input. We note that \citet{fu2023bpnet} originally used the meshes in the ABC dataset~\cite{Koch_2019_CVPR} and had to recover the patch information and degrees with a heuristic~\cite[Section 4.1]{fu2023bpnet}; by using our library, this information is readily available as it maintains the B-reps and directly samples the parametric surfaces. We selected 1\,000 random assembly parts from the Assembly dataset and sampled and labeled them with 6,000 points.
\Cref{tab:segmentiation} shows that the results of the model using \textsc{ABS} on a different dataset are consistent with the data reported by \citet{fu2023bpnet}.

\begin{lstlisting}[language=Python,frame=lines,  caption=Getting normals and primitive degrees., label=lst:segmentation]
def find_primitive_degrees(part, topo, points):
    if not topo.is_face(): return None
    
    normal = topo.normal(points)
    shape_name = topo.surface.shape_name

    if shape_name == 'BSpline': 
       if topo.surface.u_rational or topo.surface.v_rational:
          return None # BPNet only labels Bezier patches
       degree = (topo.surface.u_degree, topo.surface.v_degree)
    elif: shape_name == 'Plane': degree = (1, 1)
    elif: shape_name == 'Sphere': degree = [(2, 2), (3, 3)] 
    else: degree [(2, 3), (3, 2)]

    return [normal, degree]
\end{lstlisting}

\begin{table}
    \centering
    \scriptsize  %
    \caption{Accuracy, primitives and times across different noise levels ($\sigma$) for recovering patch degrees using BPNet.}
    \label{tab:segmentiation}
    \begin{tabular}{lccc}
        \toprule
        \textbf{Noise Level ($\sigma$)} & \textbf{Accuracy} & \textbf{Number of Primitives} & \textbf{Inference Time} \\
        \midrule
        $\sigma=0$      &        85.78 \%     &             23         &       1.63       \\
        $\sigma=0.05$      &       85.59 \%          &         24           &     1.53             \\
        $\sigma=0.1$              &      83.89 \%         &       27               &    1.65            \\
        \bottomrule
    \end{tabular}
\end{table}

\section{Conclusion}

We introduced a new open, cross-platform, and cross-language format equivalent to a B-rep, along with a Python library based on OpenCascade to convert STEP files, and a library to process the resulting format. We hope our format and library will become the new standard representation for CAD processing and machine learning on parametric surfaces. We envision pipelines where our format serves as a bridge between CAD software and state-of-the-art research.

While we have already converted several million models, more datasets remain, and we hope that the community will join the effort. Additionally, our conversion algorithm is based on OpenCascade, the only open-source CAD kernel; however, the format itself does not depend on it. Since different STEP files require different kernels, we believe our set of tools can be extended to support other (including commercial) CAD kernels.

\begin{ack}

\end{ack}

\bibliographystyle{plainnat}
\bibliography{99-biblio}

\appendix
\section{File format}\label{app:format}

The root of the HDF5 file includes one unique group called parts and has one string attribute \texttt{version} (currently version 2.0). The part group contains as many sub-groups as the model has parts, called \texttt{part\_<n>}. Each  \texttt{part\_<n>} group contains \emph{three} groups: \texttt{geometry}, \texttt{topology}, and \texttt{mesh}.

\subsection{Geometry.} 

Geometry contains the list of 2D/3D curves, the surfaces, the dataset of vertices and the bounding box of the model.

\paragraph{Curves.} Each curve can be either a circle $C$, an ellipse $E$, a line $L$, a b-spline, or an Other. All curves contain the \texttt{type} (a string encoding the name), an \texttt{interval} (the parametric space), and a \texttt{transform} (encoded as a $3\times4$ matrix in homogenous coordinates) for 3d curves. The parameterization for a curve in $\RR^N$ are
\begin{equation*}
\begin{split}
L(t)&=l + t \, d \\
C(t)&=l + r (\cos(t) a_x + \sin(t) a_y)\\
E(t)&=(f_1+f_2)/2 + r_M \cos(t) a_x + r_m \sin(t) a_y,
\end{split}
\end{equation*}
where $l\in\RR^N$ is the \texttt{location}, 
$d\in\RR^N$ the \texttt{direction}, $r\in\RR$ the \texttt{radius},
$a_x\in\RR^N$ the \texttt{x\_axis},  $a_y\in\RR^N$ the \texttt{y\_axis}, $f_1\in\RR^N$ the \texttt{focus1}, $f_2\in\RR^N$ the \texttt{focus2}, $r_M\in\RR^N$ the \texttt{maj\_radius}, and $r_m\in\RR^N$ the \texttt{min\_radius}. For a b-spline, we store the \texttt{poles} (control points) and \texttt{knots}; if it is \texttt{rational}, we have the  \texttt{weights}. We also track if the curve is \texttt{periodic} or if it  \texttt{closed}. Finally, we keep track of the \texttt{degree} and the \texttt{continuity} of the curve.

\paragraph{Surfaces} Surfaces can be either a Plane $P$, Cylinder $C_y$, Cone $C_n$, Sphere $S$, Torus $T$, BSpline, Extrusion, Revolution, or Offset. All surfaces contain \texttt{trim\_domain} (the two-dimensional parametric domain), a \texttt{transform}, and a \texttt{type}. The parameterizations are
\begin{equation*}
\begin{split}
P(u,v) &= l + u a_x + v  a_y \\
C_y(u,v) &= l + r  \cos(u)  a_x + r  \sin(u) a_y + v a_z \\
C_n(u,v) &= l + (r + v \sin(\alpha)) (\cos(u) a_x + \sin(u) a_y) + v  \cos(\alpha)  a_z\\
S(u,v) &= l + r \cos(v) (\cos(u) a_x + \sin(u)  a_y) r  \sin(v) a_z \\
T(u,v) &= l + (r_M + r_m \cos(v)) (\cos(u) a_x + \sin(u)  a_y) + r_m \sin(v) a_z,
\end{split}
\end{equation*}
where $l\in\RR^3$ is the \texttt{location}, $a_x\in\RR^3$ the \texttt{x\_axis},  $a_y\in\RR^3$ the \texttt{y\_axis}, $a_z\in\RR^3$ the \texttt{z\_axis}, $r\in\RR$ the \texttt{radius}, $\alpha\in\RR$ the \texttt{angle}, $r_M\in\RR^N$ the \texttt{max\_radius}, and $r_m\in\RR^N$ the \texttt{min\_radius}. For a b-spline, we store the \texttt{poles} (control points), \texttt{u\_knots} and  \texttt{v\_knots}; if it is \texttt{u\_rational} or  \texttt{v\_rational}, we have the  \texttt{weights}. We also track if the curve is \texttt{u\_periodic}/\texttt{v\_periodic} or if it is \texttt{u\_closed}/\texttt{v\_closed}. Finally, we keep track of the \texttt{u\_degree}, \texttt{v\_degree}, and the \texttt{continuity} of the curve.

Extrusion $E$ and Revolution $R$ contain a parametric \texttt{curve} $\gamma$ following the same standard curve definition.
\begin{equation*}
\begin{split}
E(u,v) &=  \gamma(u) + v d\\
R(u, v) &= \mathcal{R}_{a}(u)(\gamma(v) - l) + l
\end{split}
\end{equation*}
where $d\in\RR^3$ is the \texttt{direction}, $l\in\RR^3$ is the \texttt{location}, and $\mathcal{R}_{a}\in\RR^{3\times3}$ is a rotation matrix round the axis \texttt{z\_axis}.

Finally, the Extrusion contains another \texttt{surface} which can be any of the surfaces and \texttt{value}. The surface is defined by extruding the point by \texttt{value} along the \texttt{surface} normal.

\subsection{Topology.} 
Topology contains 6 groups: \texttt{edges}, \texttt{faces}, \texttt{halfedges}, \texttt{loops}, \texttt{shells}, and \texttt{solids}. All groups contain numerical subgroups, one for every entity. For instance, \texttt{/parts/part\_001/topology/solids/001} represents the second solid for the first part and \texttt{/parts/part\_001/topology/halfedges/003} the fourth half-edge.

\paragraph{Solids.} Each numerical subgroup represents one per solid in the model, each storing one dataset \texttt{shells} containing the shell indices. Note that some models have no solid as they are made of only shells; in that case, the solid group has no sub-groups. 

\paragraph{Shells.} Each shell has two datasets: \texttt{faces} and \texttt{orientation\_wrt\_solid}; the faces contain face indices, and the orientation boolean flag is used to determine the orientation of the shell. If the flag is false, the orientation of the shell must be flipped.

\paragraph{Faces.} Each face has \texttt{exact\_domain}, \texttt{has\_singularities}, \texttt{loops}, \texttt{nr\_singularities}, \texttt{outer\_loop}, \texttt{singularities}, \texttt{surface}, and \texttt{surface\_orientation}. 
\texttt{Exact\_domain} has the exact UV bounds of all loops on the face. The loops contain the indices of the loops in the face, and the outer loop is the index of the loop that contains all other loops. We also record the number of singularities (if any) and their location in the singularities group. The surface containing the index of the geometric parametric surface attached to this face. Similarly to the shells, the orientation of the face is decided by the orientation flag.

\paragraph{Loops.} Each loop contain one unique dataset \texttt{halfedges} containing indices to the half-edges.

\paragraph{Half-edge.} Every half-hedge has \texttt{2dcurve}, \texttt{edge}, \texttt{mates}, and \texttt{orientation\_wrt\_edge}. The 2d curve is an index for the \emph{geometric} 2d curve, while the edge and mates points to the \emph{topological} edges. Since multiple loops might share edges, the orientation flag indicates if the curve requires flipping.

\paragraph{Edge.} The edge is the leaf of the tree that contains only pointers to the geometry: \texttt{3dcurve} to a 3d curve, and \texttt{start\_vertex} and \texttt{end\_vertex} to vertices.

\subsection{Mesh.}

The mesh group is divided into numerical subgroups, one for each face, with each subgroup storing two datasets: \texttt{points} and \texttt{triangles}, which define the mesh. For instance, \texttt{/parts/part\_001/mesh/003/points} and \texttt{/parts/part\_001/mesh/003/triangle} contain the mesh for the fourth patch of the first part. If a face has no mesh, the corresponding \texttt{points} and \texttt{triangles} datasets will be empty.

\end{document}